\documentclass[lettersize,journal]{IEEEtran}
\usepackage{amsmath,amsfonts}
\usepackage{algorithmic}
\usepackage{algorithm}
\usepackage{array}
\usepackage{textcomp}
\usepackage{stfloats}
\usepackage{url}
\usepackage{verbatim}
\usepackage{subfigure}
\usepackage{graphicx}
\usepackage{cite}
\usepackage{multirow}
\usepackage{color}
\usepackage[table]{xcolor}
\usepackage{wrapfig}
\usepackage{pgfplots} 
\begin{document}

\title{LeNo: Adversarial Robust Salient Object Detection Networks with Learnable Noise}
\author{He Wang\textsuperscript{\rm 1,\rm 2}, Lin Wan\textsuperscript{\rm 1}, He Tang\textsuperscript{\rm 1}\thanks{The corresponding author is He Tang.}\\
~\IEEEmembership{\textsuperscript{\rm 1}School of Software Engineering, Huazhong University of Science and Technology\\
    \textsuperscript{\rm 2}School of Cyber Science and Engineering, Huazhong University of Science and Technology\\
}}



\maketitle

\begin{abstract}
Pixel-wise prediction with deep neural network has become an effective paradigm for salient object detection (SOD) and achieved remarkable performance. However, very few SOD models are robust against adversarial attacks which are visually imperceptible for human visual attention. The previous work robust saliency (ROSA) shuffles the pre-segmented superpixels and then refines the coarse saliency map by the densely connected conditional random field (CRF). Different from ROSA that relies on various pre- and post-processings, this paper proposes a light-weight Learnable Noise (LeNo) to defend adversarial attacks for SOD models. LeNo preserves accuracy of SOD models on both adversarial and clean images, as well as inference speed. In general, LeNo consists of a simple shallow noise and noise estimation that embedded in the encoder and decoder of arbitrary SOD networks respectively. Inspired by the center prior of human visual attention mechanism, we initialize the shallow noise with a cross-shaped gaussian distribution for better defense against adversarial attacks. Instead of adding additional network components for post-processing, the proposed noise estimation modifies only one channel of the decoder. With the deeply-supervised noise-decoupled training on state-of-the-art RGB and RGB-D SOD networks, LeNo outperforms previous works not only on adversarial images but also on clean images, which contributes stronger robustness for SOD. Our code is available at \url{https://github.com/ssecv/LeNo}.
\end{abstract}

\section{Introduction}
\begin{figure}[!ht]
\includegraphics[scale=0.25]{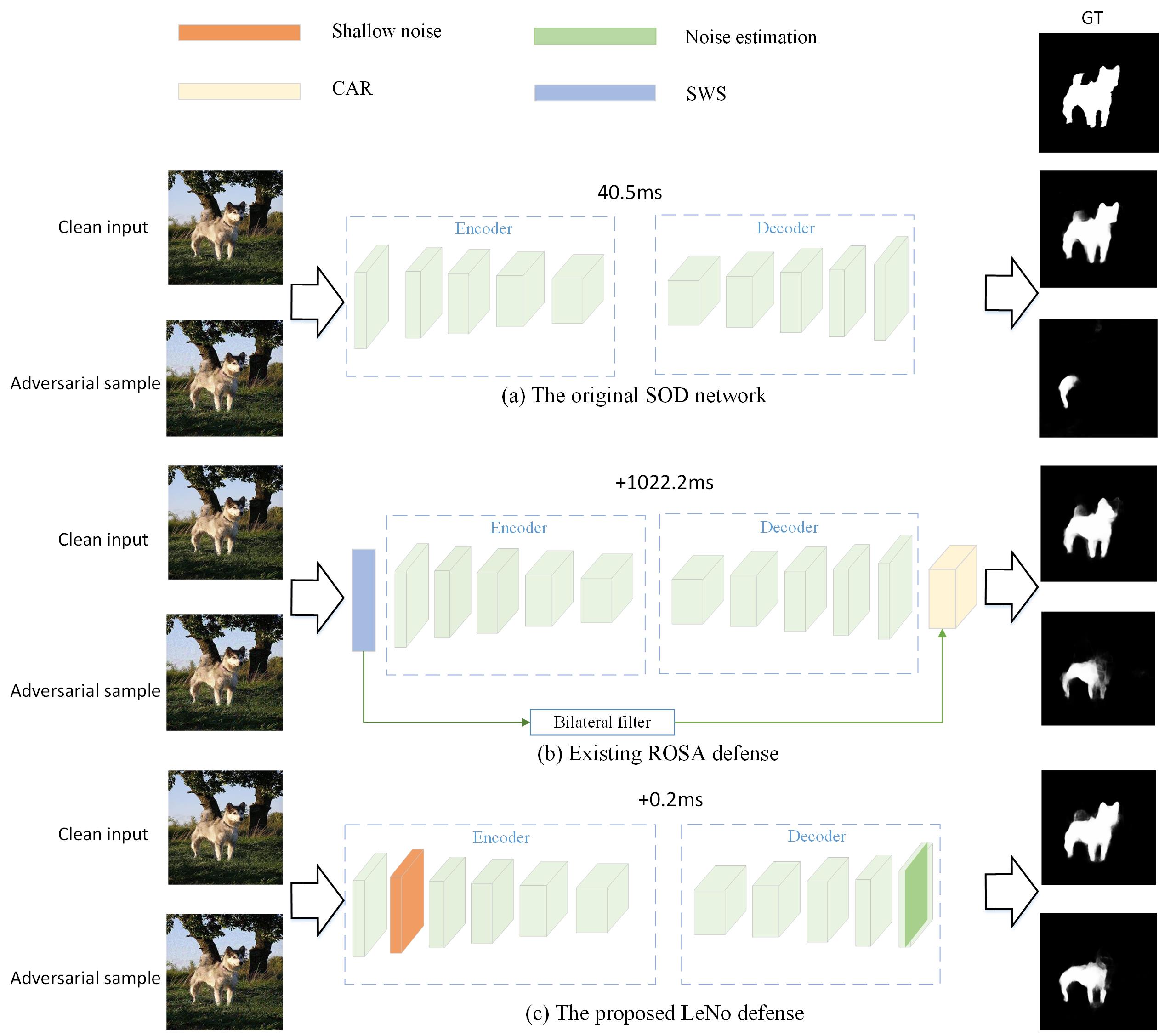}
\caption{SOD adversarial defense comparison. (a) Original GateNet without any defense. (b) ROSA defense  which adds three components to the front, middle and back of the network respectively. (c) The proposed LeNo defense, it embeds a lightweight learnable shallow noise and noise estimation and balances the performance of the clean image and adversarial image.}
\label{fig: using form}
\end{figure}

The progresses of deep neural networks (DNNs) have significantly promoted the development of down-stream computer vision tasks such as image recognition \cite{resnet}, semantic segmentation \cite{chen2017deeplab}, object detection \cite{ren2015faster} and salient object detection \cite{gatenet}. These data-driven models are usually trained with extensive input images and the annotations. Previous stuides \cite{che2021adversarial} and \cite{rosa} have shown that the saliency detection networks are fragile to adversarial attacks and the performance significantly decreases with even imperceptible perturbations. As shown in Fig. 1(a), the SOD network fails to detect the salient object of input image with adversarial perturbation, even if the salient object of the adversarial image is still obvious and can be easily distinguished from the background. As shown in Table 1, without modifying the ground truth annotations, under adversarial attacks, GateNet \cite{gatenet} obtains only .2741 $F_{\beta}$ on ECSSD dataset and BBSNet \cite{bbs} obtains only .3397 $F_{\beta}$ on NJU2K dataset, the $F_{\beta}$ drops by .6697 and .5677 respectively. This phenomenon indicates that state-of-the-art SOD models are easy to be fooled by adversarial attacks, even with the clean auxiliary depth map. The robustness of SOD models seriously concerns the security and perception gap between human and neural networks, because human visual system is hard to recognize the adversarial perturbations of the input image.

As shown in Fig. 1(b), recent study ROSA \cite{rosa} has proposed a robust salient object detection framework that incorporates a segment-wise shielding (SWS) component in front of the backbone, a context-aware restoration (CAR) component after the decoder and a bilateral filter between input image and densely connected CRF. It performs well on 4 RGB SOD benchmark datasets with 3 SOD networks. The core idea of ROSA is to introduce another generic noise to destroy subtle curve-like pattern of adversarial perturbations. SWS component divides an image into superpixels by SLIC \cite{slic} and shuffles the pixels within a superpixel randomly, it breaks adversarial pertubation by introducing random noise. The CAR component refines the saliency detection with a densely connected CRFasRNN \cite{crfasrnn}. However, the noise introduced by SWS is random and not learnable, making accuracy of the SOD model drop obviously on clean images, e.g., Fig. 1(b). Moreover, the ROSA is heavy and requires over 1 second additional time at inference stage. Thus it may retard some real-time SOD models.

In order to handle the aforementioned limitations, this paper proposes a learnable noise against adversarial attacks of SOD networks. The learnable noise consists of a shallow noise and a noise estimation. Different from SWS that introduces noise directly to the input image, the shallow noise inserts a noisy layer between stem and stage 1 of the backbone. It introduces noise in feature-level, thereby the noise is learnable and able to balance between learning clean images and adversarial images. Inspired from the image denoising method \cite{cbdnet}, we propose a lightweight noise estimation component to refine the feature of adversarial images. Our shallow noise and noise estimation are embedded in the encoder and decoder respectively, allowing parallel computation. Furthermore, the noise estimation only affects one channel of the decoder. Consequently, our defense method introduces much less extra time and performs better than ROSA, see Fig. 1(b) and (c).

Our main contributions can be summarized as three-folds:
\begin{enumerate}
  \item We launch adversarial attacks on both state-of-the-art RGB and RGB-D SOD models successfully. Experimental results verify that a wide range of existing SOD models are sensitive to adversarial perturbations.
  \item We propose a simple but efficient learnable noise (LeNo) which hardly modifies the original SOD network structure. It consists of a plug-and-play shallow noise and noise estimation. It is parallel computing and hardly influences the inference speed.
  \item With the deeply-supervised noise-decoupled training scheme, the proposed defense method promotes adversarial robustness of extensive RGB and RGB-D SOD networks. The experimental results show that our proposed defense method outperforms previous works not only on adversarial images but also clean images.
\end{enumerate}

\section{Related Works}
\subsection{Salient Object Detection}
An impressive mechanism of human vision system is the internal process that quickly scans the global image to obtain region of interest. In the field of computer vision, this task is referred to as Salient Object Detection. It plays a key role in a range of real-world applications, such as medical image segmentation \cite{12,18}, camouflaged object detection \cite{19}, etc. Although significant progress has been made in the past several years \cite{21,22,23}, there is still room for improvement when faced with challenging factors, such as complicated backgrounds or varying lighting conditions in the scenes. One way to overcome such challenges is to employ depth maps, which provides complementary spatial information and have become easier to capture due to the ready availability of depth sensors. Recently, RGB-D based salient object detection has gained increasing attention, and various methods have been developed \cite{24,25}. Early RGB-D based salient object detection models tended to extract handcrafted features and then fused the RGB image and depth map. Despite the effectiveness of traditional methods using handcrafted features, their low-level features tend to limit generalization ability, and they lack the high-level reasoning required for complex scenes. To address these limitations, several deep learning based RGB-D salient object detection methods \cite{27} have been developed, with improved performance.

\subsection{Adversarial Attacks}
Existing adversarial attacks consist of several groups, one-step gradient-based methods; iterative methods \cite{38}; optimization-based methods \cite{46}; and generative networks \cite{48,47} based methods.

\subsubsection{FGSM}
Fast Gradient Sign Method (FGSM) \cite{36} is an efficient single-step adversarial attack method. Given vectorized input $x$ and corresponding target label $y$, FGSM alters each element of $x$ along the direction of its gradient w.r.t the inference loss ${\partial L}/{\partial x}$. The generation of adversarial example $\hat{x}$ (i.e., perturbed input) can be described as:
\begin{small}
\begin{equation}
\hat{x}=x+\epsilon \cdot sgn \left(\nabla_ {x} \mathcal{L}(g(x;\theta), y)\right),
\end{equation}
\end{small}where $\epsilon$ is the perturbation constraint that determines the attack strength. $g(x;\theta)$ computes the output of DNN paramterized by $\theta$. $sgn\left(\cdot \right)$ is the sign function.

\subsubsection{PGD}
Projected Gradient Descent (PGD) \cite{30} is a multi-step variant of FGSM, which is one of the strongest $L_{\infty}$ adversarial example generation algorithm. With $\hat{x}_{t=1}=x$ as the initialization, the iterative update of perturbed data $\hat{x}$ in iteration t can be expressed as:
\begin{small}
\begin{equation}
\hat{x}_{t}=\Pi_{P_{\epsilon}(x)}\left(\hat{x}_{t-1}+a \cdot sgn\left(\nabla_{x} \mathcal{L}\left(g\left(\hat{x}_{t-1} ; \theta\right), y\right)\right)\right),
\end{equation}
\end{small}where $P_{\epsilon}(x)$ is the projection space which is bounded by $x \pm \epsilon$, and $a$ is the step size. \cite{30} also propose that PGD is a universal adversary among all the first-order adversaries.

\subsubsection{ROSA}
ROSA \cite{rosa} is an iterative gradient-based pipeline. It is the first adversarial attack on the state-of-the-art salient object detection models. They try to make the predictions of all pixels in $x$ go wrong. In each iteration t, supposing that adversarial sample $\hat{x}$ from previous time step or initialization is prepared, the adversarial sample is updated as:
\begin{small}
\begin{equation}
\hat{x}_{0}=x, \hat{x}_{t+1}=\hat{x}_{t}+p_{t},
\end{equation}
\end{small}
\begin{small}
\begin{equation}
p_{t}^{\prime}=\sum_{i \in S_{t}}\left[\nabla \hat{x}_{t} g_{i, 1-y_{i}}\left(\hat{x}_{t} ; \theta\right)-\nabla \hat{x}_{t} g_{i, y_{i}}\left(\hat{x}_{t} ; \theta\right)\right].
\end{equation}
\end{small}Here, $p_{t}$ denotes the adversarial perturbation computed at t-th step, it is obtained by normalization as $\alpha \cdot p_{t}^{\prime}/\left\|p_{t}^{\prime}\right\|_{\infty}$ where $\alpha$ is a fixed step length, i denotes one pixel in $x$, $S_{t}$ denotes the set of pixels that $g$ can still classify correctly and $y_{i}$ denotes two categories: salient and nonsalient.

\subsection{Defenses Against Adversarial Attacks}
Many researchers resort to randomization schemes \cite{random2,random3} for mitigating the effects of adversarial perturbations in the input/feature domain. The intuition behind this type of defense is that DNNs are always robust to random perturbations. A randomization based defense attempts to randomize the adversarial effects into random effects, which are not a concern for most DNNs. Some of them also add noise to the network as we do, but their noise is random and not learnable, like ROSA.

Previous works in feature denoising \cite{fdenoise} attempts to alleviate the effects of adversarial perturbations on high-level features learned by DNNs. Unlike them, we defend against adversarial perturbations by affecting low-level features, making a lightweight defense framework.

L2P \cite{l2p} is an end-to-end feature perturbation learning approach for image classification. Inspired by it, we adopt alternating back-propagation to effectively update parameters. The difference is that we propose a new cross-shaped noise with fewer parameters and without adversarial training.

\section{Methodology}
\subsection{Adversarial Attacks on Salient Object Detection Models}
We carry out a total of three attacks FGSM, PGD and ROSA. Their step sizes are 0.3, 0.04 and 0.1, respectively. The max interations of PGD and ROSA are chosen as 10 and 30. Our bound is set to 20 just like ROSA. For RGB-D salient object detection, we generate the corresponding adversarial dataset for every dataset used in our experiment with pretrained BBSNet. These adv-datasets are used to attack other RGB-D models. In this process, we include white-box attack for BBSNet and black-box attack for CoNet and MobileSal. For RGB salient object detection, same as before, we choose GateNet as the victim model. PiCANet and PFSNet are tested on adv-datasets generated by GateNet.

\subsection{Learnable Noise Against Adversarial Attacks}
\begin{figure}[!ht]
\centering
\includegraphics[scale=0.6]{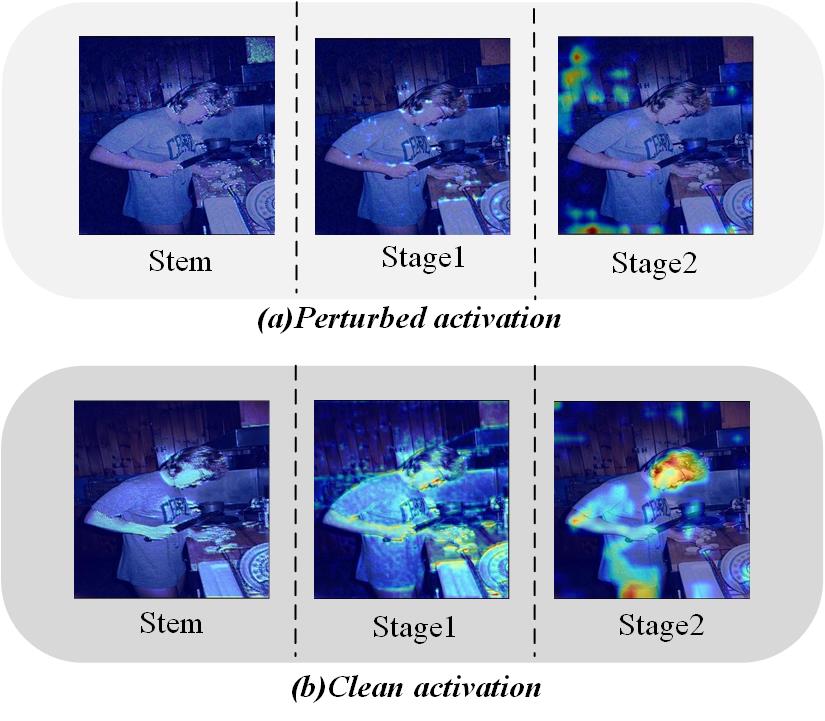}
\caption{Activation maps when PGD perturbed image and clean image are passed through backbone.}
\label{fig:1}
\end{figure}

SOD is a pixel-wise dense prediction task. The receptive field of shallow layer is smaller than deep layer and could be sensitive to low-level perturbation, such as adversarial noise. As shown in Fig. \ref{fig:1}, adversarial images cause a great suppression of the shallow activation maps of the network, the few remaining activation lead to the final huge deviation in deep layers. In this paper, our goal is to rectify the perturbated activations as much as possible. Besides, high-level features are aggregated from low-level features in DNN, so it is more effective to defend after the stem than at deeper stages. This is the core motivation for proposing the shallow noise.

\begin{figure}[!ht]
\includegraphics[scale=0.32]{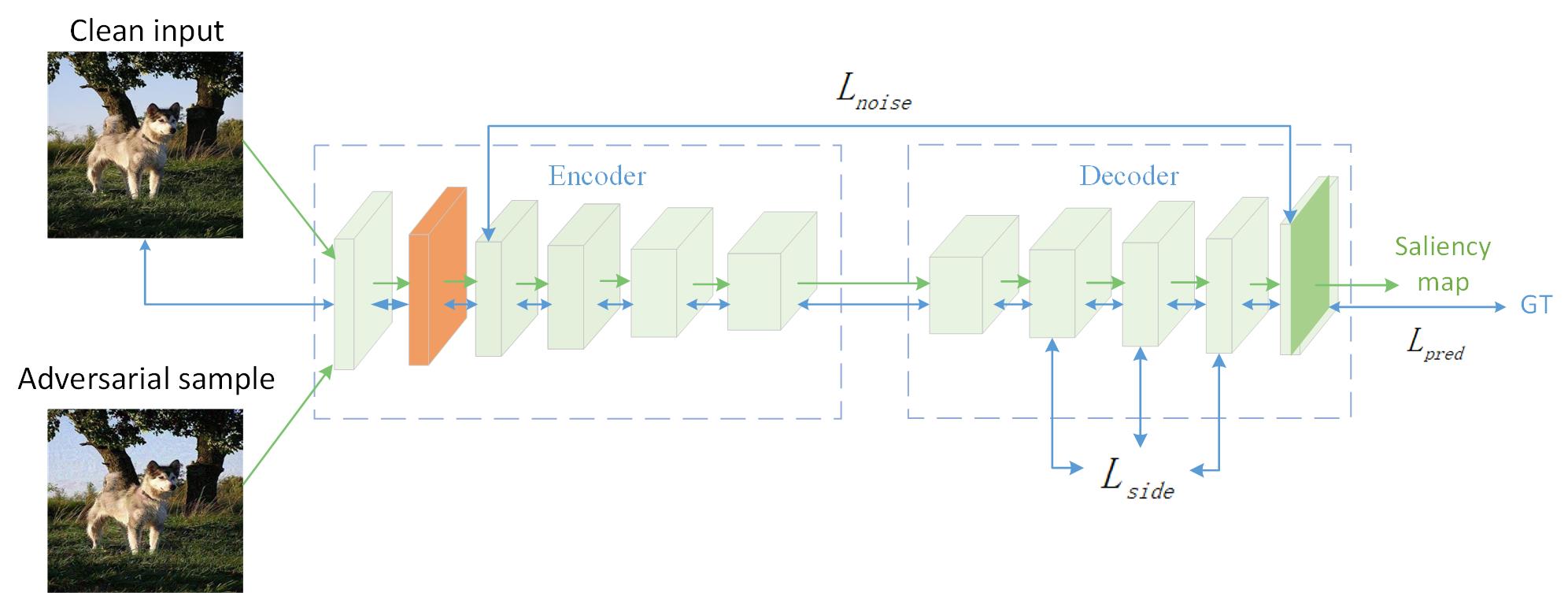}
\caption{The overall architecture of LeNo. The shallow noise is inserted into the backbone and noise estimation is part of the last stage of decoder. The blue arrows are the data flow during training. The output of shallow noise is used to calculate loss of noise estimation, i,e. $L_{noise}$. The green arrows are the data flow when LeNo detects salient objects.}
\label{fig: archi}
\end{figure}

\begin{figure}[!ht]
\includegraphics[scale=0.2]{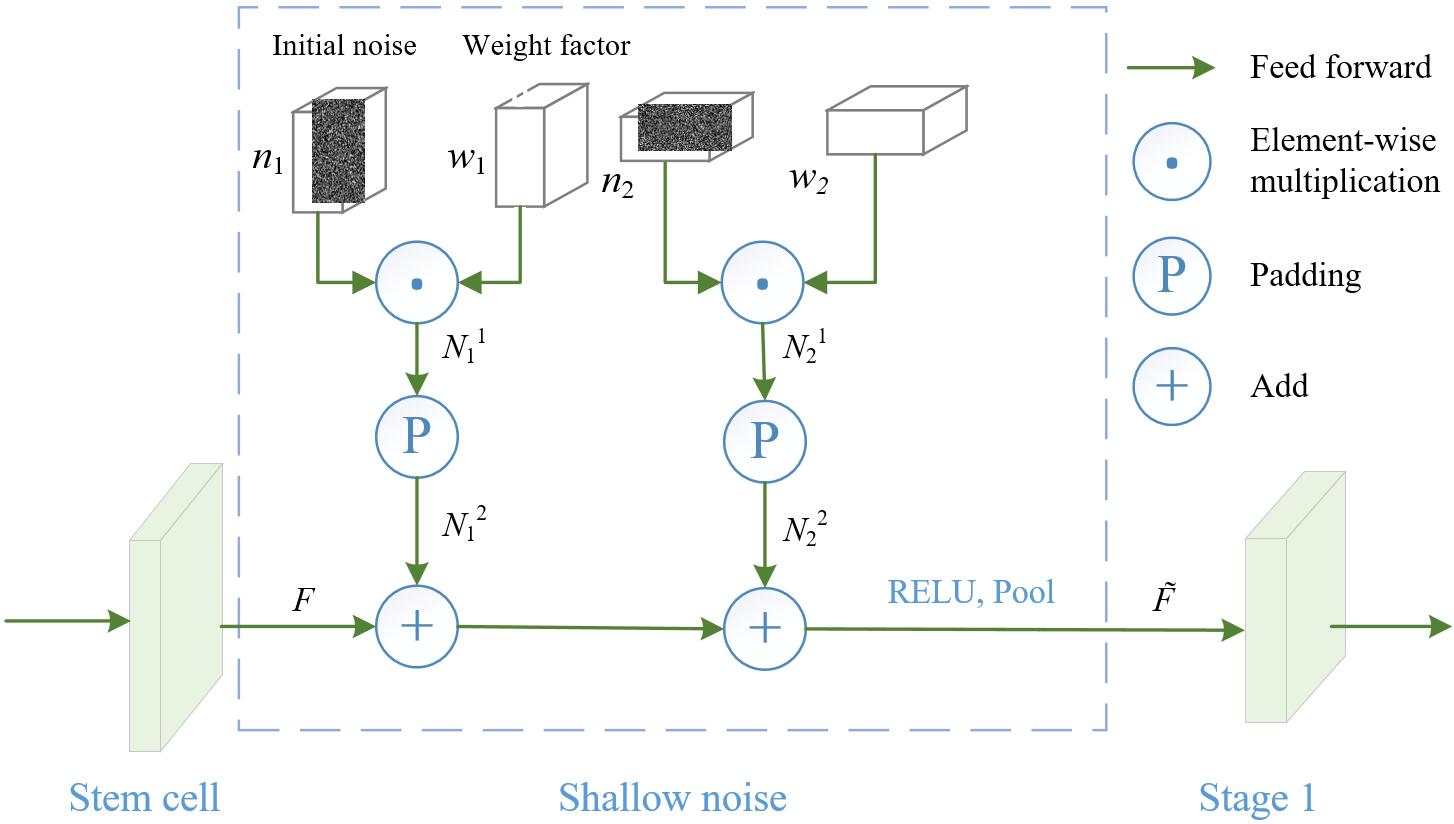}
\centering
\caption{Illustration of shallow noise.}
\label{fig: pipeline}
\end{figure}

\textbf{Shallow Noise}. We attempt to rectify the perturbed activation by adding noise to the shallow stage, as shown in Fig. \ref{fig: archi} and Fig. \ref{fig: pipeline}. First, we initialize two noises, which are shaped as $\mathcal{R}^{C\times H\times (W/2)}$ and $\mathcal{R}^{C\times (H/2)\times W}$. Here, $\mathcal{R}^{C\times H\times W}$ is the shape of input of shallow noise. Both of them are sampled from gaussian distribution and multiplied with a weight factor independently. The weight factor has the same shape with its corresponding noise and each element of it is initialized as $0.25$. The whole process can be mathematically described as:
\begin{equation}
    N_{1}^{1}=n_{1}*w_{1}, n_{1} \sim N(0,1),
\end{equation}
\begin{equation}
    N_{2}^{1}=n_{2}*w_{2}, n_{2} \sim N(0,1).
\end{equation}
Then, we pad them with zeros to shape of $\mathcal{R}^{C\times H\times W}$ as:
\begin{equation}
    N_{1}^{2}= padding(N_{1}^{1}),
\end{equation}
\begin{equation}
    N_{2}^{2}=padding(N_{2}^{1}).
\end{equation}
At this point, the noise is generated and can be added directly to the input. The output of shallow noise can be mathematically represented as:
\begin{equation}
    \tilde{F}=maxpool(RELU(F+N_{1}^{2}+N_{2}^{2})),
\end{equation}
where $F$ is the original noise-free feature which is appended with learnable noise.

\begin{figure}[!ht]
\includegraphics[scale=0.4]{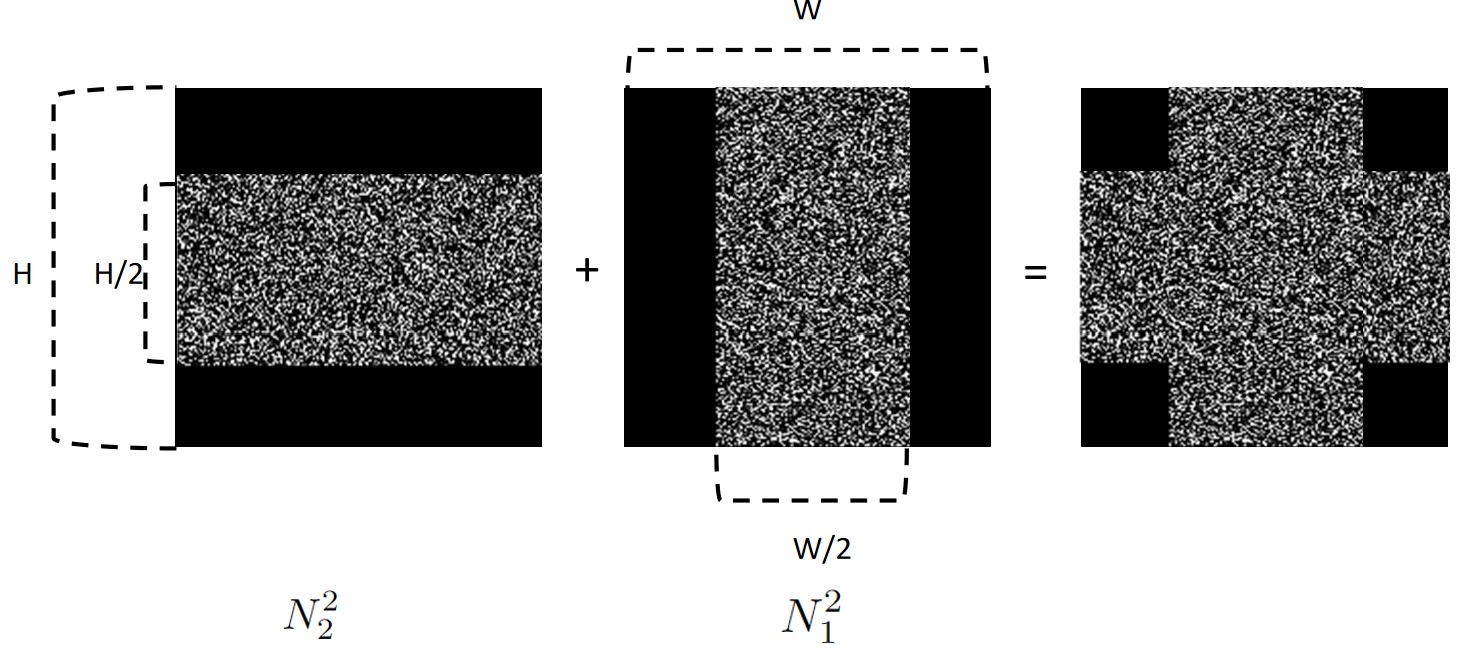}
\caption{Generation of cross-shaped noise.}
\label{figs:2}
\end{figure}

In each channel of $N_{1}^{2}$ and $N_{2}^{2}$, noise is not global. Some area has noise which is added later while others don't. As shown in Fig. \ref{figs:2}, inspired by the center prior of human visual attention mechanism, we propose a cross-shaped noise which is stronger at the center area and noise-free in the corner.

\textbf{Noise Estimation}. We replace a channel of the last stage of decoder to estimate the noise and refine the feature, minimizing changes to the network while improving the predictive performance. At the last stage of the decoder, we select the first channel for this task. We calculate the average map of output of the shallow noise along channel dimension as noise ground truth to supervise the learning process. It can be mathematically expressed as:
\begin{small}
\begin{equation}
    N_{gt}=Avg(\tilde{F}),
\end{equation}
\end{small}where $\tilde{F}$ is the output of shallow noise. Please see more details about noise estimation in the supplementary material.

\subsection{Deeply-supervised Noise-decoupled Training Scheme}
As shown in Fig. \ref{fig: archi}, the network is trained with only clean images, but performs well on both clean and adversarial images. The loss function can be mathematically discribed as:
\begin{small}
\begin{equation}
    L_{noise}=BCE(N_{est},N_{gt})+\lambda MSE(N_{est},N_{gt}),
\end{equation}
\end{small}
\begin{small}
\begin{equation}
    L_{side}=\sum_{i=1}^{3} BCE(S_{i},G),
\end{equation}
\end{small}
\begin{small}
\begin{equation}
    L_{pred}=BCE(S,G),
\end{equation}
\end{small}
\begin{small}
\begin{equation}
    L=L_{noise}+L_{side}+L_{pred},
\end{equation}
\end{small}where $N_{est}$ denotes noise estimation result in decoder, $N_{gt}$ denotes noise GT which is the average map of the output of shallow noise on channels. $S_{i}$ denotes one of the sideouts induced from decoder. S is the output of the SOD network. $L$ is the summation of the three losses, we set $\lambda$ as 0.1 in our experiments.

Broadly speaking, our training process is divided into two phases. In the first phase, we decouple the learning process to effectively update network and noise parameters. In the second phase, we freeze parameters of shallow noise and previous layers to get noise GT, preparing for noise estimation. Specifically, our detailed training procedure is described as follows:
\begin{enumerate}
    \item Initialize the network parameters $\theta_{n}^{0}$ with the pretrained weight.
    \item Randomly sampling shallow noise from a Gaussian distribution, with weights $\theta_{w}^{0}$ initialized as 0.25 for all elements.
    \item Based on $\theta_{n}^{0}$, $\theta_{w}^{0}$ and the sum of $L_{side}$ and $L_{pred}$, utilize SGD to train the network and alternately update these two parameters, obtaining $\theta_{n}^{1}$ and $\theta_{w}^{1}$. 
    \item Freeze $\theta_{w}^{1}$ and parameters before shallow noise. 
    \item Based on $\theta_{n}^{1}$, $\theta_{w}^{1}$ and $L$, utilize SGD to train the network and update network parameters, obtaining $\theta_{n}^{2}$. 
\end{enumerate}
\begin{table*}[htbp]
\caption{RGB and RGB-D SOD models under adversarial attacks.}
\tiny
\centering
\resizebox{\textwidth}{!}{
\begin{tabular}{c|cccc|cccc|cccc}
\hline
     & \multicolumn{4}{c|}{NJU2K}                       & \multicolumn{4}{c|}{NLPR}                        & \multicolumn{4}{c}{SIP}                       \\RGB-D Model  
                             & \multicolumn{2}{c}{$F_{\beta}\uparrow$} & \multicolumn{2}{c|}{MAE$\downarrow$} & \multicolumn{2}{c}{$F_{\beta}\uparrow$} & \multicolumn{2}{c|}{MAE$\downarrow$} & \multicolumn{2}{c}{$F_{\beta}\uparrow$} & \multicolumn{2}{c}{MAE$\downarrow$} \\ \cline{2-13}
                             & clean    & adver     & clean      & adver      & clean    & adver     & clean      & adver      & clean    & adver     & clean     & adver      \\ \hline
BBSNet                          & .9074   & .3397   & .0419      & .3783     & .8991   & .1976   & .0286      & .4492     & .8780   & .2718   & .0589     & .3755     \\
CoNet                        & .8946   & .7952   & .0455      & .0885     & .8928   & .7736   & .0280       & .0626     & .8693   & .7981   & .0607     & .0883     \\
MobileSal                    & .9356   & .7711   & .0312      & .1174     & .8902   & .7081   & .0336      & .0956     & .8640   & .7149  & .0703     & .1236     \\ \hline
    & \multicolumn{4}{c|}{ECSSD}                       & \multicolumn{4}{c|}{HKU-IS}                      & \multicolumn{4}{c}{DUTS-TE}                     \\RGB Model
                             & \multicolumn{2}{c}{$F_{\beta}\uparrow$} & \multicolumn{2}{c|}{MAE$\downarrow$} & \multicolumn{2}{c}{$F_{\beta}\uparrow$} & \multicolumn{2}{c|}{MAE$\downarrow$} & \multicolumn{2}{c}{$F_{\beta}\uparrow$} & \multicolumn{2}{c}{MAE$\downarrow$} \\ \cline{2-13}
                             & clean    & adver     & clean      & adver      & clean    & adver     & clean      & adver      & clean    & adver     & clean     & adver      \\ \hline
GateNet                      & .9438   & .2741   & .0332      & .4591     & .9432   & .3127   & .0301      & .4293     & .8920   & .2077   & .0382     & .4547     \\
PiCANet                      & .8770   & .6846   & .0582      & .1273     & .8710   & .7275   & .0524      & .0995     & .7984   & .5916   & .0540      & .1119     \\
PFSNet                       & .9461   & .5938   & .0305      & .2075     & .9434   & .6941   & .0285      & .1625     & .8913   & .5186   & .0366     & .2134     \\
EDNet                        & .9476   & .4351   & .0265      & .2847     & .9431   & .5357   & .0245      & .2412     & .8913   & .3708   & .0341     & .2851     \\ \hline
\end{tabular}}
\label{tab:1}
\end{table*}

\section{Experimental Results}
\subsection{Datasets}
In this paper, we focus on both RGB-D and RGB based salient object detection. For RGB-D based salient object detection, we experiment on NJU2K \cite{nju2k}, NLPR \cite{nlpr}, LFSD \cite{lfsd}, SIP \cite{sip}, NJUD \cite{njud}, STEREO \cite{stereo} and DUTS-D \cite{dutsd}. We train on the training set of NJU2K and NLPR, their test set and other public datasets are all for testing. For RGB salient object detection, we experiment on ECSSD \cite{ecssd}, HKU-IS \cite{hkuis}, DUTS \cite{dut}, DUT-OMRON \cite{duto}, PASCAL-S \cite{pascals} and SOD \cite{sod}. We train on DUTS-TR, the remaining datasets are used for testing. We follow the released data split.

\subsection{Evaluation Metrics}
We select MAE and $F_{\beta}$-measure as evaluation metrics. MAE measures pixel-level difference between the saliency map S and ground truth G as
\begin{small}
\begin{equation}
\mathrm{MAE}=\frac{1}{W \times H} \sum_{i=1}^{W} \sum_{j=1}^{H}\left|S_{i, j}-G_{i, j}\right|,
\end{equation}
\end{small}where $W$ and $H$ denote the width and height of the saliency map, respectively. Precision is the ratio of ground truth
salient pixels in the predicted salient area while recall is the ratio of predicted salient pixels in the ground truth salient area. $F_{\beta}$-measure is defined as
\begin{small}
\begin{equation}
F_{\beta}=\frac{\left(1+\beta^{2}\right) \times Precision  \times Recall }{\beta \times Precision + Recall },
\end{equation}
\end{small}where $\beta^2$ is set as 0.3 to emphasize the precision.

\subsection{Sensitivity of the Models to Adversarial Attacks}
 We demonstrate the performance of three RGB-D based visual saliency models: 1) BBSNet \cite{bbs}; 2) CoNet \cite{conet}; 3) MobileSal \cite{ms} on natural images and PGD adversarial samples which are synthesized with a pretrained BBSNet. For RGB based visual saliency models, we demonstrate the performance of: 1) GateNet \cite{gatenet}; 2) PiCANet \cite{picanet}; 3) PFSNet \cite{pfs} 4) EDNet \cite{edn} on natural images and PGD adversarial samples which are synthesized with a pretrained GateNet.
 
As shown in Table \ref{tab:1}, $F_{\beta}$-measure of BBSNet and GateNet drop .56–.7 when exposed to the adversarial samples. And $F_{\beta}$-measure of EDNet drops .41–.52 in adversarial environment. The adversarial attack reduces $F_{\beta}$-measure of PFSNet by .24–.37 while it lowers $F_{\beta}$ of CoNet, MobileSal and PiCANet by .07–.2. As shown in Table \ref{tab:1}, MAE of BBSNet and GateNet are increased by .3578 and .4183, respectively, on the adversarial samples. MAE of EDNet are raised by around .251 while that of PFSNet are raised by around .1626. MAE of CoNet, MobileSal and PiCANet change around .05. These results indicate that BBSNet and GateNet suffer most from the adversarial attack since the adversarial samples are synthesized with their own architecture. CoNet, MobileSal, PiCANet, PFSNet and EDNet are affected to different extent, which may depend on the similarity between their architectures and the model used to launch attacks. Overall, adversarial attack has a non-negligible impact on all experimental models.

\subsection{Adversarial Defense Compare With the State-of-the-art}
To demonstrate the performance of our method, we present experiments with six state-of-the-art saliency models. Our base models include BBSNet, CoNet, MobileSal, GateNet, PiCANet and PFSNet which cover RGB-D based and RGB based salient object detection. We compare our method with existing defending algorithm ROSA \cite{rosa} and L2P \cite{l2p}. ROSA is the most related work which is the first and up to date adversarial defense designed for SOD without adversarial training. We do not experiment on the base models used in ROSA to abtain up to date results. We treat ROSA as a defense method rather than a specific network and conduct experiments on a wide range of models. In all of the above settings, our approach is \textbf{universally applicable}.

\begin{figure}[!ht]
\centering
\includegraphics[scale=0.48]{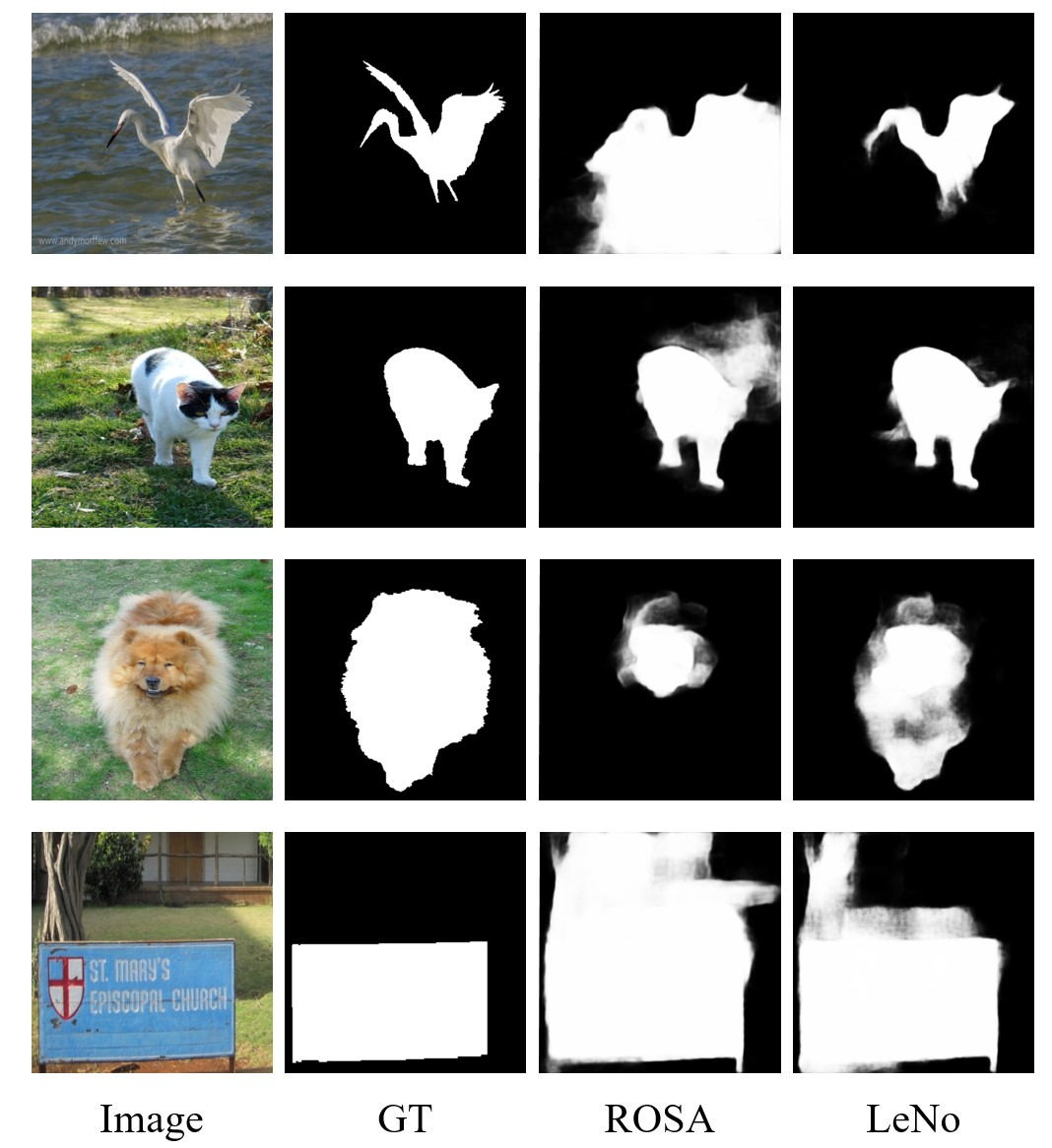}
\caption{Qualitative comparison with the state-of-the-art. The leftmost column is the adversarial images. The second column is the ground truth. The third, the fourth and the last column denote GateNet prediction incorporated with ROSA, L2P and LeNo defense respectively.}
\label{fig:quali}
\end{figure}

As Table \ref{tab:2} and Tabel \ref{tab:4} show, our method achieves better $F_{\beta}$ and MAE in most environments. For clean images, LeNo perfroms at most .0328 $F_{\beta}$ higher and .0276 MAE lower than second place. For adversarial images, our proposed method is at most .0358 $F_{\beta}$ higher and .0199 MAE lower than second place. Here, dataset-adv denotes experimenting on the adversarial samples of corresponding dataset. As Table \ref{tab:time} shows, our method has almost no effect on inference time compared with ROSA. Qualitative results are shown in Figure \ref{fig:quali}, all the input images are adversarial samples. The visual comparison demonstrates that our proposed LeNo strengthen the robustness of SOD network, the final saliency maps are closer to GT than others. Please see the supplementary materials for more quantitative results of RGB-D SOD models.

\begin{table*}[htbp]
\renewcommand\arraystretch{1.1}
\tiny
\caption{Comparison with the state-of-the-art for RGB based models on clean images}
\centering
\resizebox{\textwidth}{!}{
\begin{tabular}{c|cc|cc|cc|cc|cc|cc}
\hline
          & \multicolumn{2}{c|}{ECSSD}       & \multicolumn{2}{c|}{HKU-IS}      & \multicolumn{2}{c|}{DUTS-TE}     & \multicolumn{2}{c|}{DUT-O}       & \multicolumn{2}{c|}{PASCAL}      & \multicolumn{2}{c}{SOD}          \\
          & $F_{\beta}\uparrow$              & MAE$\downarrow$             & $F_{\beta}\uparrow$              & MAE$\downarrow$             & $F_{\beta}\uparrow$              & MAE$\downarrow$             & $F_{\beta}\uparrow$              & MAE$\downarrow$             & $F_{\beta}\uparrow$              & MAE$\downarrow$             & $F_{\beta}\uparrow$              & MAE$\downarrow$             \\ \hline
GateNet   & .9438          & .0332          & .9432          & .0301          & .8920          & .0382          & .8190           & .0545          & .8750           & .0653          & .8714          & .0990           \\
GN+ROSA   & .9292          & .0425          & .9275          & .0377          & .8592          & .0486          & .7969          & .0614          & \textbf{.8592} & \textbf{.0767} & \textbf{.8572} & \textbf{.1090}  \\
GN+L2P    & .8355          & .1097          & .8244          & .0986          & .7063          & .1185          & .6850           & .1190           & .7661          & .1436          & .7402          & .1774          \\
GN+LeNo   & \textbf{.9360} & \textbf{.0399} & \textbf{.9279} & \textbf{.0372} & \textbf{.8612} & \textbf{.0471} & \textbf{.7975} & \textbf{.0606} & .8514          & .0776          & .8462          & .1116          \\ \hline
PiCANet   & .9278          & .0455          & .9214          & .0419          & .8455          & .0525          & .7762          & .0661          & .8490          & .0785          & .8426          & .1082          \\
PN+ROSA   & .9022          & .0536          & .9073          & .0514          & .8216          & .0609          & .7659          & .0737          & .8340           & .0905          & .8272          & .1172          \\
PN+L2P    & .9177          & .0490           & .9157          & .0472          & \textbf{.8338} & \textbf{.0575} & .7741          & .0727          & \textbf{.8380}  & \textbf{.0868} & .8306          & .1163          \\
PN+LeNo   & \textbf{.9264} & \textbf{.0482} & \textbf{.9183} & \textbf{.0464} & .8316          & .0597          & \textbf{.7768} & \textbf{.0704} & .8347          & .0891          & \textbf{.8337} & \textbf{.1145} \\ \hline
PFSNet    & .9461          & .0305          & .9434          & .0285          & .8913          & .0366          & .8197          & .0553          & .8740           & .0637          & .8815          & .0890           \\
PFSN+ROSA & .8778          & .0698          & .8943          & .0491          & .8193          & .0592          & .7725          & \textbf{.0615} & .8101          & .0982          & .8122          & .1386          \\
PFSN+L2P  & .9222          & .0525          & .9031          & .0496          & .8242          & .0620           & .7751          & .0683          & .8194          & .0988          & .8190           & .1365          \\
PFSN+LeNo & \textbf{.9263} & \textbf{.0391} & \textbf{.9249} & \textbf{.0387} & \textbf{.8525} & \textbf{.0521} & \textbf{.7929} & .0656          & \textbf{.8373} & \textbf{.0840}  & \textbf{.8518} & \textbf{.1089} \\ \hline
\end{tabular}}
\label{tab:2}
\end{table*}

\begin{table*}[htbp]
\renewcommand\arraystretch{1.3}
\caption{Comparison with the state-of-the-art for RGB based models on adversarial images}
\small
\centering
\resizebox{\textwidth}{!}{
\begin{tabular}{cc|cccc|cccc|cccc}
\hline
\multicolumn{2}{c|}{}                 & \multicolumn{4}{c|}{PGD ATTACK}                         & \multicolumn{4}{c|}{ROSA ATTACK}                              & \multicolumn{4}{c}{FGSM ATTACK}                         \\ \hline
Dataset                      & Metric & GateNet & GN+ROSA   & GN+L2P          & GN+LeNo         & GateNet & GN+ROSA         & GN+L2P          & GN+LeNo         & GateNet & GN+ROSA   & GN+L2P          & GN+LeNo         \\
HKU-IS-adv  & $F_{\beta}\uparrow$      & .3127   & .5544     & .6515           & \textbf{.6538}  & .7998   & .9210            & .8171           & \textbf{.9262}  & .7720    & .8314     & .7762           & \textbf{.8379}  \\
                             & MAE$\downarrow$    & .4293  & .2532    & .1857          & \textbf{.1842} & .1118  & .0420           & .1009          & \textbf{.0394} & .0967  & .0769    & .1142          & \textbf{.0739} \\
DUTS-TE-adv & $F_{\beta}\uparrow$      & .2077   & .3757     & .4934           & \textbf{.5019}  & .7909   & .8603           & .7230         & \textbf{.8606}  & .6503   & .7095     & .6534           & \textbf{.7284}  \\
                             & MAE$\downarrow$    & .4547  & .3184    & .2450           & \textbf{.2285} & .0909  & .0501          & .1194          & \textbf{.0479} & .1123  & .1016    & .1306          & \textbf{.0936} \\
DUT-O-adv   & $F_{\beta}\uparrow$      & .1848   & .2930     & .4055           & \textbf{.4170}   & .6825   & .7935           & .6801        & \textbf{.7981}  & .5808   & .6273     & .6393           & \textbf{.6442}  \\
                             & MAE$\downarrow$    & .4882  & .3647    & .2760           & \textbf{.2725} & .1002  & .0594          & .1207          & \textbf{.0578} & .1249  & .1215    & .1294          & \textbf{.1160}  \\
PASCAL-adv  & $F_{\beta}\uparrow$      & .2915   & .4645     & \textbf{.5997}  & .5697           & .7480    & \textbf{.8505}  & .7635         & .8498           & .6798   & .7361     & .7268           & \textbf{.7514}  \\
                             & MAE$\downarrow$    & .4696  & .3300      & \textbf{.1761} & .2509          & .1608  & .0855          & .1451          & \textbf{.0821} & .1603  & .1375    & .1629          & \textbf{.1293} \\ \hline
Dataset                      & Metric & PiCANet & PN+ROSA   & PN+L2P          & PN+LeNo         & PiCANet & PN+ROSA         & PN+L2P          & PN+LeNo         & PiCANet & PN+ROSA   & PN+L2P          & PN+LeNo         \\
HKU-IS-adv  & $F_{\beta}\uparrow$      & .8081   & .8324     & .8343           & \textbf{.8468}  & .9172   & .9031           & .9129           & \textbf{.9133}  & .8498   & .8604     & .8635           & \textbf{.8712}  \\
                             & MAE$\downarrow$    & .0892  & .0819    & .0811          & \textbf{.0782} & .0443  & .0522          & .0497          & \textbf{.0488} & .0670   & .0668    & .0638          & \textbf{.0635} \\
DUTS-TE-adv & $F_{\beta}\uparrow$      & .6690    & .6971     & .7055           & \textbf{.7188}  & .8415   & .8172           & .8277           & \textbf{.8279}  & .7431   & .7501     & .7594           & \textbf{.7629}  \\
                             & MAE$\downarrow$    & .1177  & .1108    & .1091          & \textbf{.1051} & .0539  & .0629          & .0615          & \textbf{.0612} & .0858  & .0843    & .0829          & \textbf{.0827} \\
DUT-O-adv   & $F_{\beta}\uparrow$      & .6167   & .6416     & .6460            & \textbf{.6658}  & .7771   & .7632           & .7725           & \textbf{.7739}  & .6879   & .7003     & .7045           & \textbf{.7102}  \\
                             & MAE$\downarrow$    & .1287  & .1269    & .1238          & \textbf{.1195} & .0659  & .0748          & \textbf{.0706} & .0731          & .0933  & .0956    & \textbf{.0937} & .0946          \\
PASCAL-adv  & $F_{\beta}\uparrow$      & .7159   & .7419     & .7416           & \textbf{.7495}  & .8420    & .8279           & \textbf{.8326}  & .8301           & .7757   & .7761     & .7865           & \textbf{.7866}  \\
                             & MAE$\downarrow$    & .1569  & .1476    & .1440           & \textbf{.1416} & .0836  & .0954          & \textbf{.0902} & .0924          & .1201  & .1204    & \textbf{.1147} & .1172          \\ \hline
Dataset                      & Metric & PFSNet  & PFSN+ROSA & PFSN+L2P        & PFSN+LeNo       & PFSNet  & PFSN+ROSA       & PFSN+L2P        & PFSN+LeNo       & PFSNet  & PFSN+ROSA & PFSN+L2P        & PFSN+LeNo       \\
HKU-IS-adv  &$F_{\beta}\uparrow$ $F_{\beta}\uparrow$     & .6941   & .7716     & .8151           & \textbf{.8291}  & .9363   & .8826           & .8962           & \textbf{.9230}   & .8447   & .8209     & .8282           & \textbf{.8640}   \\
                             & MAE$\downarrow$    & .1625  & .0997    & .0871          & \textbf{.0809} & .0344  & .0560           & .0530           & \textbf{.0398} & .0681  & .0777    & .0827          & \textbf{.0628} \\
DUTS-TE-adv & $F_{\beta}\uparrow$      & .5186   & .6168     & .6927           & \textbf{.6992}  & .8886   & .8053           & .8149           & \textbf{.8480}   & .7347   & .6969     & .7205           & \textbf{.7541}  \\
                             & MAE$\downarrow$    & .2134  & .1292    & .1203          & \textbf{.1146} & .0395  & .0652          & .0644          & \textbf{.0544} & .0863  & .0933    & .0943          & \textbf{.0852} \\
DUT-O-adv   & $F_{\beta}\uparrow$      & .4571   & .5879     & .6383           & \textbf{.6511}  & .8205   & .7652           & .7665           & \textbf{.7951}  & .6700     & .6685     & .6987           & \textbf{.7036}  \\
                             & MAE$\downarrow$    & .2417  & .1318    & .1295          & \textbf{.1284} & .0515  & \textbf{.0639} & .0697          & .0655          & .0985  & .0909    & \textbf{.0887} & .0941          \\
PASCAL-adv  & $F_{\beta}\uparrow$      & .5998   & .6564     & \textbf{.7221}  & .7143           & .8611   & .7944           & .8162           & \textbf{.8369}  & .7512   & .7253     & .7483           & \textbf{.7612}  \\
                             & MAE$\downarrow$    & .2247  & .1779    & \textbf{.1543} & .1611          & .0717  & .1077          & .1019          & \textbf{.0878} & .1259  & .1404    & .1423          & \textbf{.1290}  \\ \hline
\end{tabular}}
\label{tab:4}
\end{table*}

\begin{table}[htbp]
\renewcommand\arraystretch{1.35}
\caption{Comparison with the state-of-the-art on speed, each inference time is averaged by 10 times computing.}
\small
\centering
\resizebox{\linewidth}{!}{
\begin{tabular}{cc|ccccccc}
\hline
BaseNet model    &  &BBS    & CoNet  & MobileSal & GateNet & PiCANet & PFSNet & EDNet  \\ \hline
     & w/o defense & 28.9   & 26.6   & 16.8      & 40.5    & 166.6   & 36.4   & 21.5   \\
 Inference    & ROSA    & +616.7 & +341.2 & +270.8    & +1022.2 & +737.2  & +550.3 & +500.2 \\
  Time(ms)    & LeNo  & \textbf{+0.5}   & \textbf{+0.2}   & \textbf{+0.1}    & \textbf{+0.2}    & \textbf{+0.1}    & \textbf{+0.3}   & \textbf{+0.3}   \\ \hline
\end{tabular}}
\label{tab:time}
\end{table}

\subsection{Ablation Studies}
\begin{table*}[htbp]
\renewcommand\arraystretch{1.2}
\caption{The ablation study of different components on the validations set of DUTS-D, NLPR, SIP and NJU2K}
\small
\centering
\resizebox{\textwidth}{!}{
\begin{tabular}{c|cccc|cccc|cccc|cccc}
\hline
      & \multicolumn{4}{c|}{DUTS-D}                       & \multicolumn{4}{c|}{NLPR}                        & \multicolumn{4}{c|}{SIP}                         & \multicolumn{4}{c}{NJU2K}                      \\
                   & \multicolumn{2}{c}{$F_{\beta}\uparrow$} & \multicolumn{2}{c|}{MAE$\downarrow$} & \multicolumn{2}{c}{$F_{\beta}\uparrow$} & \multicolumn{2}{c|}{MAE$\downarrow$} & \multicolumn{2}{c}{$F_{\beta}\uparrow$} & \multicolumn{2}{c|}{MAE$\downarrow$} & \multicolumn{2}{c}{$F_{\beta}\uparrow$} & \multicolumn{2}{c}{MAE$\downarrow$} \\
                   & clean     & adver     & clean       & adver      & clean     & adver     & clean       & adver      & clean     & adver     & clean       & adver      & clean     & adver     & clean      & adver      \\ \hline
BBSNet (base model) & .8117   & .2473   & .0828      & .5398     & .8991   & .1976   & .0286      & .4492     & .8780   & .2718   & .0589      & .3755     & .9074   & .3397   & .0419     & .3783     \\\hline
+Shallow noise     & .7594   & .5197   & .1059      & .2382     & .8535   & .7580   & .0413      & .0767     & .8599   & .7395   & .0689      & .1184     & .8831   & .7582   & .0539     & .1157     \\
+Noise estimation   & .7852   & .5125   & .0977     & .2413     & .8744   & .7574   & .0385      & .0819     & .8628   & .7317   & .0681      & .1245     & .8868   & .7531   & .0536     & .1191     \\ \hline
\end{tabular}}
\label{tab:components}
\end{table*}

\begin{table}[htbp]
\renewcommand\arraystretch{1.3}
\caption{Effect of different noise initial distribution}
\tiny
\centering
\resizebox{\linewidth}{!}{
\begin{tabular}{ccccccc}
\hline
Dataset & \multicolumn{2}{c}{NJU2K-adv}   & \multicolumn{2}{c}{SIP-adv}   & \multicolumn{2}{c}{DUTS-D-adv} \\ \hline
Metric           & $F_{\beta}\uparrow$                    & MAE$\downarrow$               & $F_{\beta}\uparrow$                  & MAE$\downarrow$               & $F_{\beta}\uparrow$                   & MAE$\downarrow$               \\
Uniform          & .7775              & \textbf{.0927}            & .6910            & .1257            & .4734             & .1899            \\
Constant         & .7652              & .0979            & .6768            & .1303            & .4922             & .1906            \\
Gaussian            & \textbf{.7876}     & .0947            & \textbf{.7252}   & \textbf{.1187}   & \textbf{.5282}      & \textbf{.1786}   \\ \hline
Dataset & \multicolumn{2}{c}{DUTS-TE-adv} & \multicolumn{2}{c}{DUT-O-adv} & \multicolumn{2}{c}{PASCAL-adv} \\ \hline
Metric           & $F_{\beta}\uparrow$                    & MAE$\downarrow$               & $F_{\beta}\uparrow$                  & MAE$\downarrow$               & $F_{\beta}\uparrow$                   & MAE$\downarrow$               \\
Uniform          & .6773              & .1078            & .6017            & .1255            & .7076             & .1491            \\
Constant         & .6754              & .1074            & .5981            & .1240             & .7039             & .1507            \\
Gaussian            & \textbf{.6950}     & \textbf{.1030}    & \textbf{.6162}   & \textbf{.1212}   & \textbf{.7201}    & \textbf{.1448}   \\ \hline
\end{tabular}}
\label{tab:initial}
\end{table}

\textbf{Contribution of each component}. As shown in Table \ref{tab:components}, after adding shallow noise to the original model, the new network's performance decreases on clean images slightly, but increases on adversarial images greatly. In the worst case, it dropped by .0523 in $F_{\beta}$ and increased by .0231 in MAE on clean images, but at this point there is a .2724$F_{\beta}$ and .3016 MAE improvement in performance against adversarial samples. After continuing to add noise estimation, the performance of the model on clean images is improved while basically maintaining the adversarial performance. 

\textbf{Noise initialization}. Three different noise initialization settings are selected. We test 1) uniform distribution with zero mean and unit variance, 2) gaussian distribution with zero mean and variance 0.734 and 3) constant setting of .5. Such a choice of parameters ensures their average perturbation intensity is equal. As Table \ref{tab:initial} shows, in most cases the difference between them is .01-.02 and Gaussian distribution is the optimal choice.

\textbf{Number of noise layer}. As Table \ref{tab:number} shows, whether on the clean images or the adversarial images, single noise layer is always the best. This is because superabundant noises may hamper the final prediction. Here, +1 denotes adding noise layer after every BottleNeck in stem and stage1. And +2 denotes adding noise layer after every BottleNeck in stem, stage1 and stage2.
\begin{table}[htbp]
\renewcommand\arraystretch{1.35}
\caption{Effect of number of noise layer}
\small
\centering
\resizebox{\linewidth}{!}{
\begin{tabular}{cl|llllll}
\hline
\multicolumn{1}{l}{\textbf{Method}} & \multicolumn{1}{c|}{\textbf{Metric}} & \multicolumn{1}{c}{\textbf{NJU2K-adv}} & \multicolumn{1}{c}{\textbf{NLPR-adv}} & \multicolumn{1}{c}{\textbf{SIP-adv}}  & \multicolumn{1}{c}{\textbf{NJU2K}} & \multicolumn{1}{c}{\textbf{NLPR}} & \multicolumn{1}{c}{\textbf{SIP}}  \\ \hline
+1        & $F_{\beta}\uparrow$                                   & .7677                                & .6805                               & \textbf{.7521}                      & .7956                            & .7056                           & .7838                           \\
                                    & MAE$\downarrow$                                 & .1104                                 & .0892                                & .1187                                & .1002                    & .0866                   & .1078                   \\
+2        & $F_{\beta}\uparrow$                                   & .7521                                & .6430                               & .7261                               & .7743                            & .6625                           & .7519                           \\
                                    & MAE$\downarrow$                                 & .1176                                 & .1017                                & .1348                                & .1124                             & .0959                            & .1255                            \\
LeNo      & $F_{\beta}\uparrow$                                   & \textbf{.7956}                       & \textbf{.7333}                      & .7345                               & \textbf{.8207}                   & \textbf{.7424}                  & \textbf{.7852}                  \\
                                    & MAE$\downarrow$                                 & \textbf{.0929}                        & \textbf{.0721}                       & \textbf{.1170}                        & \textbf{.0815}                    & \textbf{.0685}                   & \textbf{.0988}                   \\ \hline
\multicolumn{1}{l}{\textbf{Method}} & \multicolumn{1}{c|}{\textbf{Metric}} & \multicolumn{1}{c}{\textbf{ECSSD-adv}} & \multicolumn{1}{c}{\textbf{HKU-adv}}  & \multicolumn{1}{c}{\textbf{DUTS-adv}} & \multicolumn{1}{c}{\textbf{ECSSD}} & \multicolumn{1}{c}{\textbf{HKU}}  & \multicolumn{1}{c}{\textbf{DUTS}} \\ \hline
+1        & $F_{\beta}\uparrow$                                   & .8310                                & .8299                               & .7249                      & .8687                            & .8678                           & .7699                           \\
                                    & MAE$\downarrow$                                 & .1062                        & .0935                                & .1015                                & .0922                    & .0826                   &.0912                   \\
+2        & $F_{\beta}\uparrow$                                   & .7993                                & .8113                               & .7053                               & .7993                            & .8113                           & .7485                  \\
                                    & MAE$\downarrow$                                 & .1200                                   & .1015                                & .1227                                & .1200                               & .1015                            & .1116                            \\
LeNo      & $F_{\beta}\uparrow$                                   & \textbf{.8775}                       & \textbf{.8712}                      & \textbf{.7629}                      & \textbf{.9264}                   & \textbf{.9183}                  & \textbf{.8316}                  \\
                                    & MAE$\downarrow$                        & \textbf{.0732}                        & \textbf{.0635}                       & \textbf{.0827}                       & \textbf{.0482}                    & \textbf{.0472}                   & \textbf{.0597}                   \\ \hline
\end{tabular}}
\label{tab:number}
\end{table}

\textbf{Noise area}. Finally, we investigated how the noise area affect the performance. We report the results in FGSM adversarial environment. Here, network-full denotes that all elements of the feature map are added with effective noise, which is adopted by \cite{l2p}. Network-center denotes only the elements in the center area are added with effective noise, which is an intuitive consideration for Salient Object Detection. Here, the central area is a square with the sides halved. Salient objects usually appear at the center part of the image, so we make the central area the focus of our defense. Adding stronger noise at the center of the feature map makes the defense more targeted. In order to balance the noise strength and computational complexity, we propose the cross-shaped noise as Fig. \ref{figs:2}. Further ablation studies are shown in the supplementary material, cross-shaped noise is the optimal.

\textbf{Inference time.} The proposed LeNo defense can be computed in parallel since it is embedded in encoder and decoder of the network. The extra time of ROSA is introduced by the summation of superpixel shuffle \cite{slic}, CRFRNN \cite{crfasrnn} and bilateral filter. As Table \ref{tab:time} shows, LeNo is significantly faster than ROSA on all base models.

\section{Conclusion}
In recent years, salient object detection with deep neural network has achieved good performance. However, it is sensitive to adversarial attacks, which rises a huge security challenge. In this paper, a lightweight Learnable Noise (LeNo) is proposed against adversarial attacks for SOD models. Different from the previous work ROSA which relies on various pre- and post-processings, the shallow noise with a cross-shaped Gaussian distribution and noise estimation for refining the feature are introduced to replace SWS and CAR components to speed up inference. Extensive experiments show that LeNo gets better result in RGB and RGB-D SOD benchmarks with much lower inference latency. \newpage

\bibliographystyle{acm}
\addcontentsline{toc}{section}{Reference}
\bibliography{arxiv}

\begin{thebibliography}{10}

\bibitem{slic}
{\sc Achanta, R., Shaji, A., Smith, K., Lucchi, A., Fua, P., and S{\"u}sstrunk,
  S.}
\newblock Slic superpixels compared to state-of-the-art superpixel methods.
\newblock {\em IEEE transactions on pattern analysis and machine intelligence
  34}, 11 (2012), 2274--2282.

\bibitem{che2021adversarial}
{\sc Che, Z., Borji, A., Zhai, G., Ling, S., Li, J., Tian, Y., Guo, G., and
  Le~Callet, P.}
\newblock Adversarial attack against deep saliency models powered by
  non-redundant priors.
\newblock {\em IEEE Transactions on Image Processing 30\/} (2021), 1973--1988.

\bibitem{25}
{\sc Chen, H., Li, Y., Deng, Y., and Lin, G.}
\newblock Cnn-based rgb-d salient object detection: Learn, select, and fuse.
\newblock {\em International Journal of Computer Vision 129}, 7 (2021),
  2076--2096.

\bibitem{chen2017deeplab}
{\sc Chen, L.-C., Papandreou, G., Kokkinos, I., Murphy, K., and Yuille, A.~L.}
\newblock Deeplab: Semantic image segmentation with deep convolutional nets,
  atrous convolution, and fully connected crfs.
\newblock {\em IEEE transactions on pattern analysis and machine intelligence
  40}, 4 (2017), 834--848.

\bibitem{random3}
{\sc Dhillon, G.~S., Azizzadenesheli, K., Lipton, Z.~C., Bernstein, J.,
  Kossaifi, J., Khanna, A., and Anandkumar, A.}
\newblock Stochastic activation pruning for robust adversarial defense.
\newblock {\em arXiv preprint arXiv:1803.01442\/} (2018).

\bibitem{38}
{\sc Dong, Y., Liao, F., Pang, T., Su, H., Zhu, J., Hu, X., and Li, J.}
\newblock Boosting adversarial attacks with momentum.
\newblock In {\em Proceedings of the IEEE conference on computer vision and
  pattern recognition\/} (2018), pp.~9185--9193.

\bibitem{19}
{\sc Fan, D.-P., Ji, G.-P., Sun, G., Cheng, M.-M., Shen, J., and Shao, L.}
\newblock Camouflaged object detection.
\newblock In {\em Proceedings of the IEEE/CVF Conference on Computer Vision and
  Pattern Recognition (CVPR)\/} (June 2020).

\bibitem{12}
{\sc Fan, D.-P., Ji, G.-P., Zhou, T., Chen, G., Fu, H., Shen, J., and Shao, L.}
\newblock Pranet: Parallel reverse attention network for polyp segmentation.
\newblock In {\em International conference on medical image computing and
  computer-assisted intervention\/} (2020), Springer, pp.~263--273.

\bibitem{24}
{\sc Fan, D.-P., Lin, Z., Zhang, Z., Zhu, M., and Cheng, M.-M.}
\newblock Rethinking rgb-d salient object detection: Models, data sets, and
  large-scale benchmarks.
\newblock {\em IEEE Transactions on neural networks and learning systems 32}, 5
  (2020), 2075--2089.

\bibitem{27}
{\sc Fan, D.-P., Lin, Z., Zhang, Z., Zhu, M., and Cheng, M.-M.}
\newblock Rethinking rgb-d salient object detection: Models, data sets, and
  large-scale benchmarks.
\newblock {\em IEEE Transactions on neural networks and learning systems 32}, 5
  (2020), 2075--2089.

\bibitem{sip}
{\sc Fan, D.-P., Lin, Z., Zhang, Z., Zhu, M., and Cheng, M.-M.}
\newblock Rethinking rgb-d salient object detection: Models, data sets, and
  large-scale benchmarks.
\newblock {\em IEEE Transactions on neural networks and learning systems 32}, 5
  (2020), 2075--2089.

\bibitem{bbs}
{\sc Fan, D.-P., Zhai, Y., Borji, A., Yang, J., and Shao, L.}
\newblock Bbs-net: Rgb-d salient object detection with a bifurcated backbone
  strategy network.
\newblock In {\em European conference on computer vision\/} (2020), Springer,
  pp.~275--292.

\bibitem{36}
{\sc Goodfellow, I.~J., Shlens, J., and Szegedy, C.}
\newblock Explaining and harnessing adversarial examples.
\newblock {\em arXiv preprint arXiv:1412.6572\/} (2014).

\bibitem{cbdnet}
{\sc Guo, S., Yan, Z., Zhang, K., Zuo, W., and Zhang, L.}
\newblock Toward convolutional blind denoising of real photographs.
\newblock In {\em Proceedings of the IEEE/CVF conference on computer vision and
  pattern recognition\/} (2019), pp.~1712--1722.

\bibitem{resnet}
{\sc He, K., Zhang, X., Ren, S., and Sun, J.}
\newblock Deep residual learning for image recognition.
\newblock In {\em Proceedings of the IEEE conference on computer vision and
  pattern recognition\/} (2016), pp.~770--778.

\bibitem{l2p}
{\sc Jeddi, A., Shafiee, M.~J., Karg, M., Scharfenberger, C., and Wong, A.}
\newblock Learn2perturb: an end-to-end feature perturbation learning to improve
  adversarial robustness.
\newblock In {\em Proceedings of the IEEE/CVF Conference on Computer Vision and
  Pattern Recognition\/} (2020), pp.~1241--1250.

\bibitem{conet}
{\sc Ji, W., Li, J., Zhang, M., Piao, Y., and Lu, H.}
\newblock Accurate rgb-d salient object detection via collaborative learning.
\newblock In {\em European Conference on Computer Vision\/} (2020), Springer,
  pp.~52--69.

\bibitem{nju2k}
{\sc Ju, R., Ge, L., Geng, W., Ren, T., and Wu, G.}
\newblock Depth saliency based on anisotropic center-surround difference.
\newblock In {\em 2014 IEEE international conference on image processing
  (ICIP)\/} (2014), IEEE, pp.~1115--1119.

\bibitem{njud}
{\sc Ju, R., Ge, L., Geng, W., Ren, T., and Wu, G.}
\newblock Depth saliency based on anisotropic center-surround difference.
\newblock In {\em 2014 IEEE international conference on image processing
  (ICIP)\/} (2014), IEEE, pp.~1115--1119.

\bibitem{hkuis}
{\sc Li, G., and Yu, Y.}
\newblock Visual saliency based on multiscale deep features.
\newblock In {\em Proceedings of the IEEE conference on computer vision and
  pattern recognition\/} (2015), pp.~5455--5463.

\bibitem{rosa}
{\sc Li, H., Li, G., and Yu, Y.}
\newblock Rosa: Robust salient object detection against adversarial attacks.
\newblock {\em IEEE transactions on cybernetics 50}, 11 (2019), 4835--4847.

\bibitem{lfsd}
{\sc Li, N., Ye, J., Ji, Y., Ling, H., and Yu, J.}
\newblock Saliency detection on light field.
\newblock In {\em Proceedings of the IEEE Conference on Computer Vision and
  Pattern Recognition\/} (2014), pp.~2806--2813.

\bibitem{22}
{\sc Li, X., Yang, F., Cheng, H., Liu, W., and Shen, D.}
\newblock Contour knowledge transfer for salient object detection.
\newblock In {\em Proceedings of the european conference on computer vision
  (ECCV)\/} (2018), pp.~355--370.

\bibitem{pascals}
{\sc Li, Y., Hou, X., Koch, C., Rehg, J.~M., and Yuille, A.~L.}
\newblock The secrets of salient object segmentation.
\newblock In {\em Proceedings of the IEEE conference on computer vision and
  pattern recognition\/} (2014), pp.~280--287.

\bibitem{fdenoise}
{\sc Liao, F., Liang, M., Dong, Y., Pang, T., Hu, X., and Zhu, J.}
\newblock Defense against adversarial attacks using high-level representation
  guided denoiser.
\newblock In {\em Proceedings of the IEEE conference on computer vision and
  pattern recognition\/} (2018), pp.~1778--1787.

\bibitem{picanet}
{\sc Liu, N., Han, J., and Yang, M.-H.}
\newblock Picanet: Learning pixel-wise contextual attention for saliency
  detection.
\newblock In {\em Proceedings of the IEEE conference on computer vision and
  pattern recognition\/} (2018), pp.~3089--3098.

\bibitem{random2}
{\sc Liu, X., Cheng, M., Zhang, H., and Hsieh, C.-J.}
\newblock Towards robust neural networks via random self-ensemble.
\newblock In {\em Proceedings of the European Conference on Computer Vision
  (ECCV)\/} (2018), pp.~369--385.

\bibitem{pfs}
{\sc Ma, M., Xia, C., and Li, J.}
\newblock Pyramidal feature shrinking for salient object detection.
\newblock In {\em Proceedings of the AAAI Conference on Artificial
  Intelligence\/} (2021), vol.~35, pp.~2311--2318.

\bibitem{30}
{\sc Madry, A., Makelov, A., Schmidt, L., Tsipras, D., and Vladu, A.}
\newblock Towards deep learning models resistant to adversarial attacks.
\newblock {\em arXiv preprint arXiv:1706.06083\/} (2017).

\bibitem{sod}
{\sc Martin, D., Fowlkes, C., Tal, D., and Malik, J.}
\newblock A database of human segmented natural images and its application to
  evaluating segmentation algorithms and measuring ecological statistics.
\newblock In {\em Proc. 8th Int'l Conf. Computer Vision\/} (July 2001), vol.~2,
  pp.~416--423.

\bibitem{stereo}
{\sc Niu, Y., Geng, Y., Li, X., and Liu, F.}
\newblock Leveraging stereopsis for saliency analysis.
\newblock In {\em 2012 IEEE Conference on Computer Vision and Pattern
  Recognition\/} (2012), IEEE, pp.~454--461.

\bibitem{nlpr}
{\sc Peng, H., Li, B., Xiong, W., Hu, W., and Ji, R.}
\newblock Rgbd salient object detection: A benchmark and algorithms.
\newblock In {\em European conference on computer vision\/} (2014), Springer,
  pp.~92--109.

\bibitem{dutsd}
{\sc Piao, Y., Ji, W., Li, J., Zhang, M., and Lu, H.}
\newblock Depth-induced multi-scale recurrent attention network for saliency
  detection.
\newblock In {\em Proceedings of the IEEE/CVF International Conference on
  Computer Vision\/} (2019), pp.~7254--7263.

\bibitem{47}
{\sc Poursaeed, O., Katsman, I., Gao, B., and Belongie, S.}
\newblock Generative adversarial perturbations.
\newblock In {\em Proceedings of the IEEE Conference on Computer Vision and
  Pattern Recognition\/} (2018), pp.~4422--4431.

\bibitem{23}
{\sc Qin, X., Zhang, Z., Huang, C., Gao, C., Dehghan, M., and Jagersand, M.}
\newblock Basnet: Boundary-aware salient object detection.
\newblock In {\em Proceedings of the IEEE/CVF conference on computer vision and
  pattern recognition\/} (2019), pp.~7479--7489.

\bibitem{ren2015faster}
{\sc Ren, S., He, K., Girshick, R., and Sun, J.}
\newblock Faster r-cnn: Towards real-time object detection with region proposal
  networks.
\newblock {\em Advances in neural information processing systems 28\/} (2015).

\bibitem{ecssd}
{\sc Shi, J., Yan, Q., Xu, L., and Jia, J.}
\newblock Hierarchical image saliency detection on extended cssd.
\newblock {\em IEEE transactions on pattern analysis and machine intelligence
  38}, 4 (2015), 717--729.

\bibitem{dut}
{\sc Wang, L., Lu, H., Wang, Y., Feng, M., Wang, D., Yin, B., and Ruan, X.}
\newblock Learning to detect salient objects with image-level supervision.
\newblock In {\em Proceedings of the IEEE conference on computer vision and
  pattern recognition\/} (2017), pp.~136--145.

\bibitem{18}
{\sc Wu, Y.-H., Gao, S.-H., Mei, J., Xu, J., Fan, D.-P., Zhang, R.-G., and
  Cheng, M.-M.}
\newblock Jcs: An explainable covid-19 diagnosis system by joint classification
  and segmentation.
\newblock {\em IEEE Transactions on Image Processing 30\/} (2021), 3113--3126.

\bibitem{ms}
{\sc Wu, Y.-H., Liu, Y., Xu, J., Bian, J.-W., Gu, Y.-C., and Cheng, M.-M.}
\newblock Mobilesal: Extremely efficient rgb-d salient object detection.
\newblock {\em IEEE Transactions on Pattern Analysis and Machine
  Intelligence\/} (2021).

\bibitem{edn}
{\sc Wu, Y.-H., Liu, Y., Zhang, L., Cheng, M.-M., and Ren, B.}
\newblock Edn: Salient object detection via extremely-downsampled network.
\newblock {\em IEEE Transactions on Image Processing 31\/} (2022), 3125--3136.

\bibitem{46}
{\sc Xiao, C., Zhu, J.-Y., Li, B., He, W., Liu, M., and Song, D.}
\newblock Spatially transformed adversarial examples.
\newblock {\em arXiv preprint arXiv:1801.02612\/} (2018).

\bibitem{duto}
{\sc Yang, C., Zhang, L., Lu, H., Ruan, X., and Yang, M.-H.}
\newblock Saliency detection via graph-based manifold ranking.
\newblock In {\em Proceedings of the IEEE conference on computer vision and
  pattern recognition\/} (2013), pp.~3166--3173.

\bibitem{21}
{\sc Zhao, J.-X., Liu, J.-J., Fan, D.-P., Cao, Y., Yang, J., and Cheng, M.-M.}
\newblock Egnet: Edge guidance network for salient object detection.
\newblock In {\em Proceedings of the IEEE/CVF international conference on
  computer vision\/} (2019), pp.~8779--8788.

\bibitem{gatenet}
{\sc Zhao, X., Pang, Y., Zhang, L., Lu, H., and Zhang, L.}
\newblock Suppress and balance: A simple gated network for salient object
  detection.
\newblock In {\em European conference on computer vision\/} (2020), Springer,
  pp.~35--51.

\bibitem{48}
{\sc Zhao, Z., Dua, D., and Singh, S.}
\newblock Generating natural adversarial examples.
\newblock {\em arXiv preprint arXiv:1710.11342\/} (2017).

\bibitem{crfasrnn}
{\sc Zheng, S., Jayasumana, S., Romera-Paredes, B., Vineet, V., Su, Z., Du, D.,
  Huang, C., and Torr, P.~H.}
\newblock Conditional random fields as recurrent neural networks.
\newblock In {\em Proceedings of the IEEE international conference on computer
  vision\/} (2015), pp.~1529--1537.

\end{thebibliography}
\end{document}